\documentclass[a4paper]{article}

\usepackage{INTERSPEECH2020}
\usepackage{multirow,CJKutf8,subfigure}
\usepackage{url,setspace}
\usepackage[breaklinks=true]{hyperref}
\usepackage[anythingbreaks]{breakurl}

\newcommand*{\Ja}[1]{\begin{CJK}{UTF8}{ipxm}#1\end{CJK}}
\newcommand{\argmin}{\mathop{\rm argmin}\limits}

\usepackage[
backend=biber,
style=ieee,
maxbibnames=5,
maxcitenames=5,
doi=false,isbn=false,url=false,eprint=false
]{biblatex} 

\defbibheading{bibliography}[\refname]{}
\DeclareSourcemap{
	\maps[datatype=bibtex, overwrite=true]{
		\map{
			\step[fieldsource=booktitle,
			match=\regexp{.*Interspeech.*},
			replace={Proc. Interspeech}]
			\step[fieldsource=journal,
			match=\regexp{.*INTERSPEECH.*},
			replace={Proc. Interspeech}]
			\step[fieldsource=booktitle,
			match=\regexp{.*ICASSP.*},
			replace={ICASSP}]
			\step[fieldsource=booktitle,
			match=\regexp{.*icassp_inpress.*},
			replace={ICASSP (in press)}]
			\step[fieldsource=booktitle,
			match=\regexp{.*International.*Conference.*on.*Acoustics,.*Speech.*and.*Signal.*Processing.*},
			replace={ICASSP}]
			\step[fieldsource=booktitle,
			match=\regexp{.*International.*Conference.*on.*Learning.*Representations.*},
			replace={ICLR}]
			\step[fieldsource=booktitle,
			match=\regexp{.*International.*Conference.*on.*Machine.*Learning.*},
			replace={ICML}]
			\step[fieldsource=booktitle,
			match=\regexp{.*Automatic.*Speech.*Recognition.*and.*Understanding.*},
			replace={Proc. ASRU}]
			\step[fieldsource=booktitle,
			match=\regexp{.*Spoken.*Language.*Technology.*},
			replace={Proc. SLT}]
			\step[fieldsource=booktitle,
			match=\regexp{.*Speech.*Synthesis.*Workshop.*},
			replace={Proc. SSW}]
			\step[fieldsource=booktitle,
			match=\regexp{.*workshop.*on.*speech.*synthesis.*},
			replace={Proc. SSW}]
			\step[fieldsource=booktitle,
			match=\regexp{.*Advances.*in.*neural.*information.*processing.*},
			replace={NIPS}]
			\step[fieldsource=booktitle,
			match=\regexp{.*Advances.*in.*Neural.*Information.*Processing.*},
			replace={NIPS}]
			\step[fieldsource=booktitle,
			match=\regexp{.*Workshop.*on.* Applications.* of.* Signal.*Processing.*to.*Audio.*and.*Acoustics.*},
			replace={Proc. WASPAA}]
			\step[fieldsource=publisher,
			match=\regexp{.+},
			replace={{}}]
			\step[fieldsource=month,
			match=\regexp{.+},
			replace={{}}]
			\step[fieldsource=location,
			match=\regexp{.+},
			replace={{}}]
			\step[fieldsource=address,
			match=\regexp{.+},
			replace={{}}]
			\step[fieldsource=organization,
			match=\regexp{.+},
			replace={{}}]
		}
	}
}

\title{DiscreTalk: Text-to-Speech as a Machine Translation Problem}
\name{Tomoki Hayashi$^{1,2}$, Shinji Watanabe$^3$}
\address{
  $^1$Human Dataware Lab. Co. Ltd., Japan\\
  $^2$Nagoya University, Japan\\
  $^3$Johns Hopkins University, USA}
\email{hayashi.tomoki@g.sp.m.is.nagoya-u.ac.jp}

\begin{document}

\maketitle
\begin{abstract}
This paper proposes a new end-to-end text-to-speech (E2E-TTS) model based on neural machine translation (NMT).
The proposed model consists of two components; a non-autoregressive vector quantized variational autoencoder (VQ-VAE) model and an autoregressive Transformer-NMT model.
The VQ-VAE model learns a mapping function from a speech waveform into a sequence of discrete symbols, and then the Transformer-NMT model is trained to estimate this discrete symbol sequence from a given input text. 
Since the VQ-VAE model can learn such a mapping in a fully-data-driven manner, we do not need to consider hyperparameters of the feature extraction required in the conventional E2E-TTS models.
Thanks to the use of discrete symbols, we can use various techniques developed in NMT and automatic speech recognition (ASR) such as beam search, subword units, and fusions with a language model.
Furthermore, we can avoid an over smoothing problem of predicted features, which is one of the common issues in TTS.
The experimental evaluation with the JSUT corpus shows that the proposed method outperforms the conventional Transformer-TTS model with a non-autoregressive neural vocoder in naturalness, achieving the performance comparable to the reconstruction of the VQ-VAE model.
\end{abstract}

\section{Introduction}
With the improvement of deep learning techniques, the presence of end-to-end text-to-speech (E2E-TTS) models has been growing not only in the research field but also as a production system~\cite{wang2017tacotron,shen2017tacotron2,ping2017deep,wei2019clarinet,li2018transformer,ren2019fastspeech}. 
Compared to the conventional statistical parametric speech synthesis (SPSS) systems~\cite{taylor2009text,zen2009statistical,tokuda2013speech}, the E2E-TTS models do not require the language expert knowledge and the alignments between text and speech, making it possible to train the system with only the pairs of text and speech.
Thanks to the neural vocoders~\cite{oord2016wavenet,wu2016investigating,tamamori2017speaker,prenger2019waveglow,Yamamoto2020,kumar2019melgan}, the E2E-TTS models can achieve the quality comparable to professionally recorded speech.

Although E2E-TTS models have achieved excellent performance with a simple training scheme, it still depends on the human-designed acoustic features, e.g., Mel-spectrogram~\cite{shen2017tacotron2} and speech parameters such as $F_0$~\cite{ping2017deep}.
To achieve the best performance, we must carefully tune the hyperparameters of these acoustic features such as the number of points in fast Fourier transform (FFT), a window size, the analysis range of frequency, and the normalization method.
Hence, current E2E-TTS models are not in a fully end-to-end manner.

Recently, the discretization of sequential information in a fully-data-driven manner by vector quantized variational autoencoder (VQ-VAE)~\cite{van2017neural} has been getting attention. 
The VQ-VAE model, which consists of encoder and decoder networks, can convert an arbitrary length sequential input into a downsampled sequence of the discrete symbols and precisely reconstruct the input from the discrete symbol sequence even if the input is a raw waveform of speech.
One of the applications using VQ-VAE for speech processing is a voice conversion, where it models speech waveforms directly and utilizes additional speaker ID embedding to condition the decoder to control the speaker characteristics~\cite{van2017neural,ding2019group}.
Tjandra {\it et al.} have extended to a more challenging task, cross-lingual voice conversion by combining with a sequence-to-sequence (Seq2Seq) model~\cite{tjandra2019speech}.
Another exciting idea is automatic speech recognition (ASR) or TTS without target text~\cite{dunbar2017zero,dunbar2019zero}, where VQ-VAE models the acoustic feature space to obtain the discretized unit similar to phoneme instead of the corresponding text~\cite{chorowski2019unsupervised,baevski2019vq}.
Henter {\it et al.} have utilized VQ-VAE to obtain the embedding to control the speech speaking style in an unsupervised manner~\cite{henter2018deep}.
Kumar {\it et al.} have applied the non-autoregressive generative adversarial network (GAN)-based VQ-VAE model for music generation~\cite{kumar2019melgan}.
Thus, VQ-VAE has excellent potential to obtain meaningful representations in an unsupervised manner even for the extreme long sequence, such as speech waveforms.

This paper proposes a novel E2E-TTS framework based on VQ-VAE and neural machine translation (NMT), which is a fully-end-to-end model without human-designed acoustic features.
The proposed model consists of two components; the VQ-VAE model that learns a mapping from a speech waveform into a sequence of discrete symbols and the Transformer-NMT model trained to estimate the discrete symbol sequence from a given input text.
Since the VQ-VAE model can learn such a mapping in a fully-data-driven manner, we do not need to consider hyperparameters of the feature extraction required in the conventional E2E-TTS models.
Thanks to the use of discrete symbols, we can use techniques developed in NMT and ASR.
Furthermore, we can avoid an over smoothing problem of the predicted features, which is one of the common issues in TTS.
The contributions of this paper are summarized as follows:
{
\setlength{\leftmargini}{12pt} 
\begin{itemize}
    \setlength{\itemsep}{-1pt}
    \item We propose a non-autoregressive GAN-based VQ-VAE model with the multi-resolution short-time Fourier transform (STFT) loss inspired by great successes of GAN-based neural vocoders~\cite{kumar2019melgan,Yamamoto2020}. The proposed VQ-VAE's decoder can decode the sequence of discrete symbols into a speech waveform much faster than the real-time while keeping the reasonable quality.
    \item We introduce advanced decoding techniques such as beam-search, shallow fusion with a language model (LM), and subword unit, commonly used in NMT and ASR fields. 
    With the ASR evaluation metric, we can investigate the effectiveness of these techniques more intuitively.
    \item The experimental results of the subjective evaluation with mean opinion score (MOS) on naturalness show that the proposed method outperforms the conventional E2E-TTS model (Transformer-TTS~\cite{li2018transformer} with Parallel WaveGAN~\cite{Yamamoto2020}),  achieving the performance comparable to the reconstruction of the VQ-VAE model.
\end{itemize}
}%

\section{VQ-VAE}
The VQ-VAE model consists of three components: an encoder, a decoder, and a shared codebook.
The encoder $\mathrm{Enc}(\cdot)$ is a non-linear mapping function to encode $T$-length arbitrary sequential input $\mathbf{X} = \{\mathbf{x}_1, \mathbf{x}_2, ..., \mathbf{x}_T\}$ into a downsampled $N$-length sequence of vectors $\mathbf{Z} = \{\mathbf{z}_1, ...,\mathbf{z}_n, ..., \mathbf{z}_N\} (N < T)$ in the latent space.
Then, the quantization function $Q(\cdot)$ converts vector $\mathbf{z}_n$ to $i$-th centroid vector $\mathbf{e}_i$ in the codebook based on the distance between them as follows:
\begin{equation}
    Q(\mathbf{z}_n) = \mathbf{z}_{\mathrm{vq}}^{(n)} = \mathbf{e}_{i} \ \  \mathrm{where} \ \ i = \argmin_{j} \|\mathbf{z}_n - \mathbf{e}_j \|.
\end{equation}
The decoder $\mathrm{Dec}(\cdot)$ is another non-linear mapping function to reconstruct the input waveform from the sequence of quantized vectors.
The whole network is trained using the following objective function:
\begin{gather}
    \label{eq:obj_fn}
    \mathcal{L} = \mathcal{L}_{\mathrm{rec}} + \mathcal{L}_{\mathrm{cb}} + \lambda_{\mathrm{cm}}\mathcal{L}_{\mathrm{cm}}, \\
    \label{eq:reconst_fn}    \mathcal{L}_{\mathrm{rec}} = \|\mathrm{Dec}(Q(\mathrm{Enc}(\mathbf{X}))) - \mathbf{X}\|^2_2, \\
    \mathcal{L}_{\mathrm{cb}} = \|\mathrm{sg}[\mathrm{Enc}(\mathbf{X})] - \mathbf{Z}_{\mathrm{vq}}\|^2_2, \\
    \mathcal{L}_{\mathrm{cm}} = \|\mathrm{sg}[\mathbf{Z}_{\mathrm{vq}}] - \mathrm{Enc}(\mathbf{X}) \|^2_2,
\end{gather}
where $\mathcal{L}_{\mathrm{rec}}$, $\mathcal{L}_{\mathrm{cb}}$, and $\mathcal{L}_{\mathrm{cm}}$ represent reconstruction loss, codebook loss, and commitment loss, respectively.
$\lambda_{\mathrm{cm}}$ represents a constant coefficient for balancing between codebook loss and commitment loss.
$\mathbf{Z}_{\mathrm{vq}}$ represents the sequence $\{\mathbf{z}_{\mathrm{vq}}^{(1)}, ..., \mathbf{z}_{\mathrm{vq}}^{(N)}\} $ and $\mathrm{sg}[\cdot]$ represents stop-gradient operation to prevent gradient from flowing its argument.

\section{Method}
\subsection{Overview}
\begin{figure}[t]
    \vspace{0mm}
    \begin{center}
        \includegraphics[width=0.95\columnwidth]{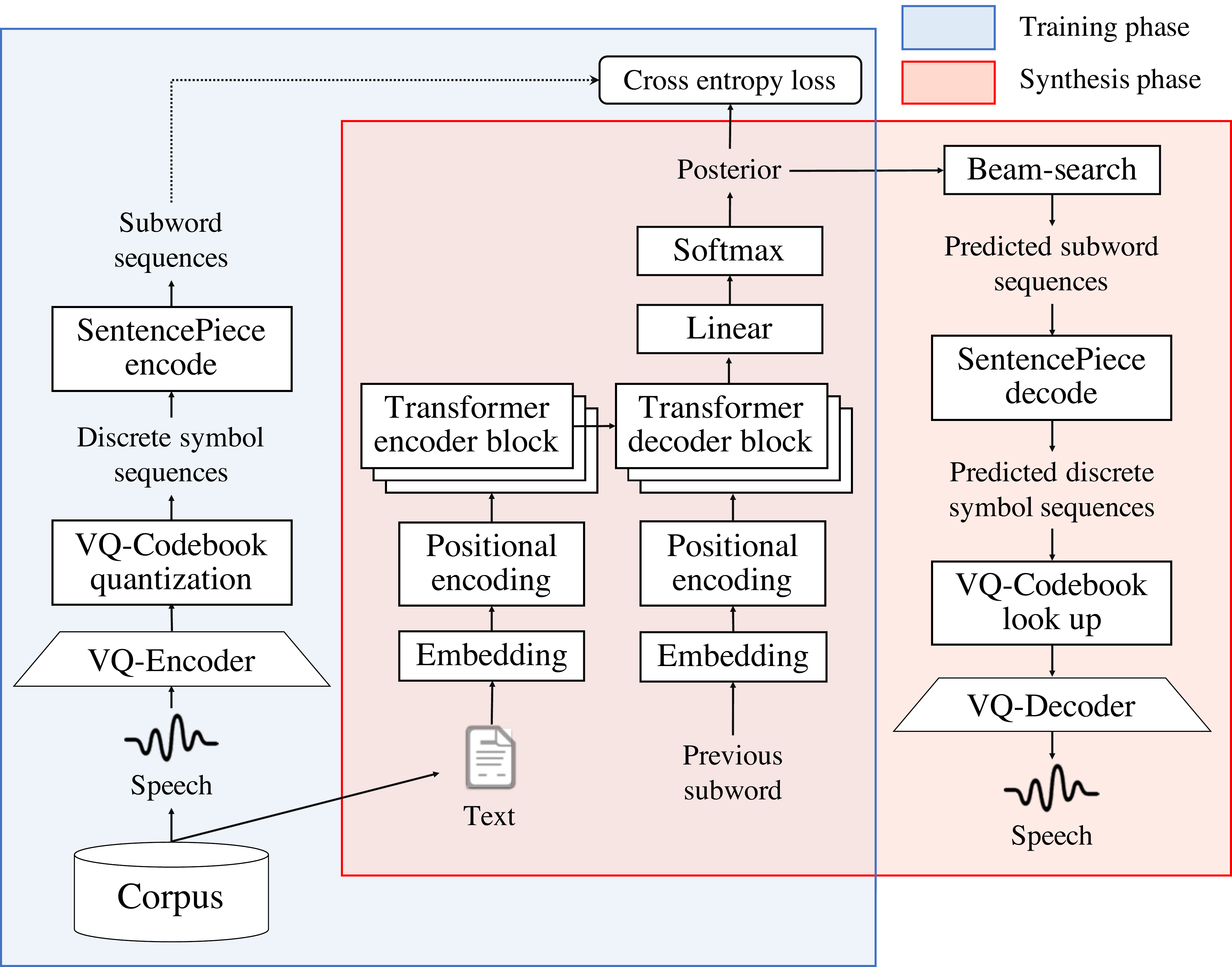}
    \end{center}
    \vspace{-5mm}
    \caption{\it Overview of the proposed method.}
    \label{fig:overview}
    \vspace{-6mm}
\end{figure}

The overview of the proposed model is shown in Fig~\ref{fig:overview}, which is divided into training and synthesis phases.
In the training phase, at first, we train the non-autoregressive GAN-based VQ-VAE model using speech waveforms in the corpus (Section \ref{sec:vq-vae}).
Next, the VQ-encoder converts all of the speech waveforms into sequences of the discrete symbols, i.e., centroid IDs in the VQ-codebook. 
The discrete symbol sequence is encoded into a sequence of subword units by SentencePiece~\cite{kudo2018sentencepiece}. 
Then, we train the Transformer-NMT model~\cite{vaswani2017attention} using text as the inputs and the subword sequence as the targets (Section \ref{sec:transformer-mt}).  

In the synthesis phase, the Transformer-NMT model estimates the subword sequence from a given text using beam-search~\cite{watanabe2017hybrid}.
The SentencePiece model decodes the estimated sequence of subword units into that of the original discrete symbols.
Then, the VQ-codebook replaces each discrete symbol with the embedding.
Finally, the VQ-decoder converts the embedding sequence into a speech waveform.
In the following sections, we explain each component in detail.

\subsection{Non-autoregressive GAN-based VQ-VAE}
\label{sec:vq-vae}
\begin{figure}[t]
    \vspace{0mm}
    \begin{center}
        \includegraphics[width=0.9\columnwidth]{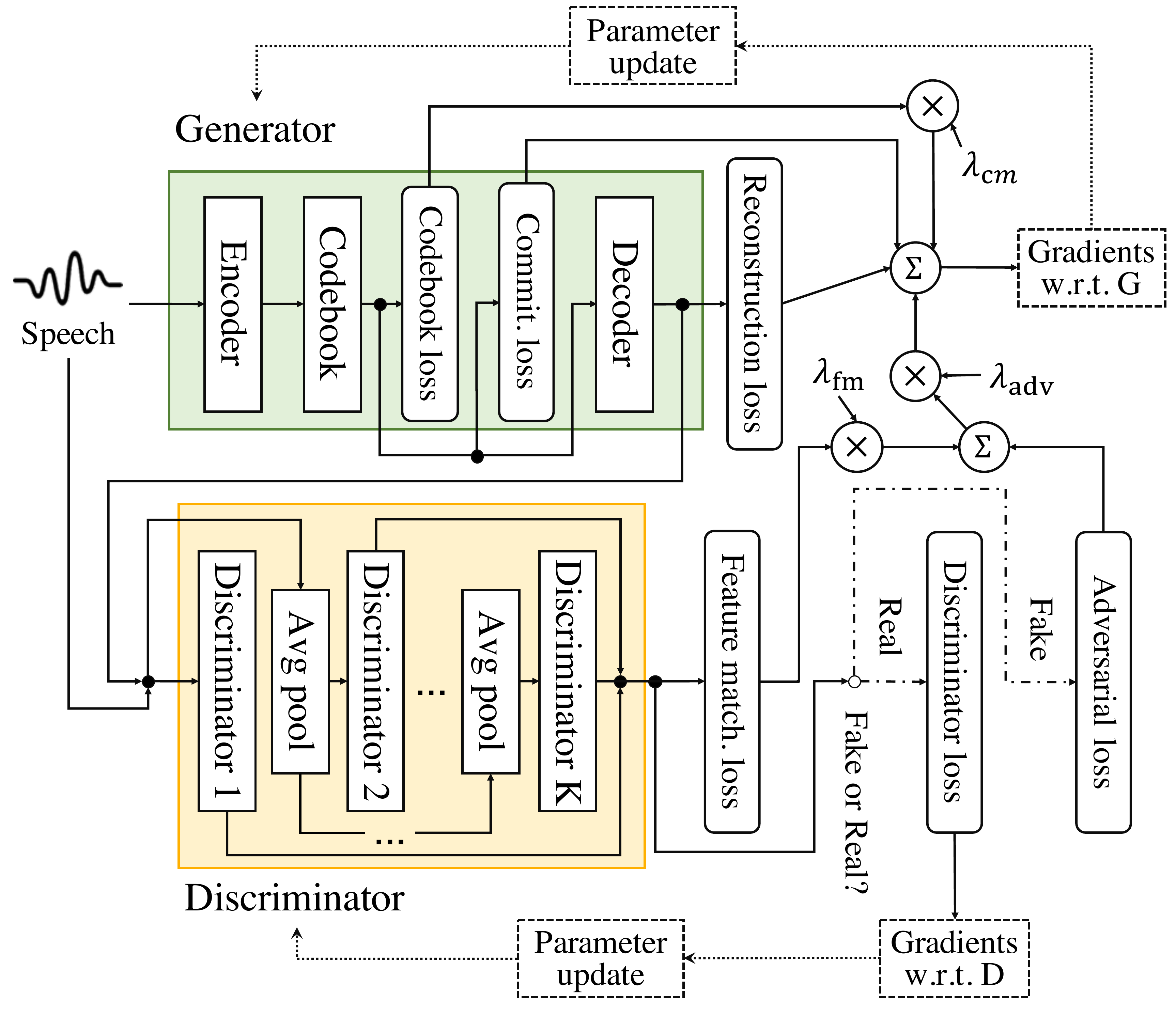}
    \end{center}
    \vspace{-6mm}
    \caption{\it The proposed VQ-VAE training flow.}
    \label{fig:vq_vae_flow}
    \vspace{-6mm}
\end{figure}
Let us introduce a speech waveform $\mathbf{x} = \{x_1, x_2, ..., x_T\}$ as the inputs of the VQ-VAE model.
To boost up the reconstruction performance, we use a multi-resolution STFT loss~\cite{Yamamoto2020} as the reconstruction function and additionally introduce an adversarial objective function based on MelGAN~\cite{kumar2019melgan}.
Let us describe the VQ-VAE model as a generator network $G(\cdot)$ ($=\mathrm{Dec}(Q(\mathrm{Enc}(\cdot)))$, and introduce $K$ discriminator networks $D_k(\cdot)\ (k=1,2,...,K)$, each of which has the same structure.
As the network structure, we use the MelGAN's generator as the decoder and the MelGAN's discriminator as the encoder and each discriminator.
The objective function in Eq.~(\ref{eq:obj_fn}) and the reconstruction function in Eq.~(\ref{eq:reconst_fn}) are modified as follows:
\begin{gather}
    \mathcal{L} = \mathcal{L}_{G} = \mathcal{L}_{\mathrm{rec}} + \mathcal{L}_{\mathrm{cb}} + \lambda_{\mathrm{cm}}\mathcal{L}_{\mathrm{cm}} + \lambda_{\mathrm{adv}}\mathcal{L}_{\mathrm{adv}}, \\
    \mathcal{L}_{\mathrm{rec}} = \dfrac{1}{M}{\scriptstyle\sum_{m=1}^M} \left(\mathcal{L}_{\mathrm{mag}}^{(m)} + \mathcal{L}_{\mathrm{sc}}^{(m)} \right), \\
    \mathcal{L}_{\mathrm{mag}}^{(m)} = \dfrac{1}{F}\|\ \log|\mathrm{STFT}(\mathbf{x})|-\log|\mathrm{STFT}(G(\mathbf{x}))|\ \|_1, \\
    \mathcal{L}_{\mathrm{sc}}^{(m)} = \dfrac{\|\ |\mathrm{STFT}(\mathbf{x})|-|\mathrm{STFT}(G(\mathbf{x}))|\ \|_F}{\|\ |\mathrm{STFT}(\mathbf{x})|\ \|_F}, \\
    \mathcal{L}_{\mathrm{adv}} = \dfrac{1}{K}{\scriptstyle\sum}_{k=1}^{K}\left( (\mathbf{1}-D_k(G(\mathbf{x})))^2 + \lambda_{\mathrm{fm}}\mathcal{L}_{\mathrm{fm}}^{(k)}\right), \\
    \mathcal{L}_{\mathrm{fm}}^{(k)} = \dfrac{1}{L}{\scriptstyle\sum}_{l=1}^{L}\| D_k^{(l)}(\mathbf{x})-D_k^{(l)}(G(\mathbf{x})) \|_1,
\end{gather}
where $\mathcal{L}_{\mathrm{sc}}$, $\mathcal{L}_{\mathrm{mag}}$, $\mathcal{L}_{\mathrm{adv}}$, and $\mathcal{L}_{\mathrm{fm}}$ represent spectral convergence loss, magnitude loss~\cite{arik2018fast}, adversarial loss and feature matching loss~\cite{kumar2019melgan}, respectively.
$M$ and $m$ represents the number of the STFT loss functions and its index, respectively.
$\|\cdot\|_1$ and $\|\cdot\|_F$ represent $L_1$ and Frobenius norm, respectively.
$\mathrm{|STFT(\cdot)|}$ and $F$ represent STFT magnitude and the number of elements in the magnitude, respectively.
${\scriptstyle {D_k^{(l)}(\cdot)}}$ and $L$ represent the $l$-th layer's outputs of $k$-th discriminator and the number of layers, respectively.
The objective function of the discriminator $\mathcal{L}_{D}$ is defined as follows:
\begin{gather}
    \
    \mathcal{L}_{D} = \dfrac{1}{K}{\scriptstyle\sum}_{k=1}^{K}\left(\mathbf{1} - D_k(\mathbf{x}) \right)^2.
\end{gather}
We summarize the flow of the training in Fig.~\ref{fig:vq_vae_flow}, which updating the generator and the discriminator alternately.


\subsection{Transformer-NMT}
\label{sec:transformer-mt}
We use an autoregressive Transformer-based encoder-decoder model~\cite{vaswani2017attention}, which consists of self-attention layers.
The model learns the mapping from the character or the phoneme sequence to that of subword units of discrete symbols converted by SentencePiece~\cite{kudo2018sentencepiece}.
Note that unlike a regular NMT problem, the length of the input and output differ significantly since the length of the output is similar to that of acoustic features. 
We use cross-entropy loss to optimize the network with the label smoothing technique~\cite{muller2019does}.

\section{Experimental Evaluation}

\subsection{Experimental condition}
To demonstrate the performance of the proposed model, we conducted an experimental evaluation using the JSUT corpus~\cite{sonobe2017jsut}.
The JSUT corpus includes 10 hours of a single female Japanese speech.
We used 7,196, 250, and 250 utterances for training, validation, and evaluation, respectively.
The speech was downsampled to 24k Hz and the input text was converted into a phoneme sequence using Open JTalk~\cite{openjtalk}.
To check the performance of the proposed method, we compared the following five models:
{
\setlength{\leftmargini}{12pt} 
\begin{enumerate}
    \setlength{\itemsep}{-1pt}
    \item {\bf Target}: The target speech. We downsampled to 24k Hz and trimmed the silence at the beginning and the end of utterances by the force alignment with Julius~\cite{lee2009recent}.
    \item {\bf Baseline}: The baseline system using Transformer-TTS~\cite{li2018transformer} with Parallel WaveGAN~\cite{Yamamoto2020}. We used the open-source toolkit ESPnet-TTS~\cite{hayashi2020espnet} to build this system.
    \item {\bf Reconst}: Reconstructed speech by the proposed VQ-VAE.
    \item {\bf Proposed~(Raw)}: The proposed model with raw discrete symbols as the target of the NMT model.
    \item {\bf Proposed~(SW)}: The proposed model with subword units.
\end{enumerate}
}
\noindent For the proposed models, we built several models using different downsampling factor (DSF) in the VQ-VAE model (DSF128 or DSF256) and the different number of subword units (SW256 or SW512)\footnote{256 is the same as the number of centroids in codebook, but that of active centroids is much smaller than predefined size.}. 
All of the generated samples are available online~\cite{audio_samples}\footnote{We are also planning to publish the codes as an open-source.}.
The detailed training condition of the VQ-VAE and Transformer-NMT models is shown in Table~\ref{tb:ex_cond_1}.
\begin{table}[t]
\begin{center}
\vspace{0mm}
\caption{\it Training conditions.}
\vspace{-2mm}
\scalebox{0.9}{\renewcommand\arraystretch{0.9}{%
\begin{tabular}{ll}
\toprule
\multicolumn{2}{l}{\bf VQ-VAE training condition}  \\
\midrule
Sampling rate & 24,000 Hz \\
\# Centroids & 256 \\
Centroids dimension & 128 \\
\multirow{2}{*}{Downsampling scales} & [4, 4, 4, 4] (for DSF256) \\
 & [4, 4, 4, 2] (for DSF128) \\
\multirow{2}{*}{Upsampling scales} & [8, 8, 2, 2] (for DSF256) \\
 & [8, 8, 4, 2] (for DSF128) \\
Batch size & 16 \\
Batch length & 8,192 \\
Optimizer & RAdam~\cite{liu2019variance} \\
\multirow{1}{*}{Learning rate} & 1e-4 (for $G$) \& 5e-5 (for $D$) \\
\multirow{1}{*}{Gradient clip} & 10.0 (for $G$) \& 1.0 (for $D$) \\
\# Iterations & 5,000,000 \\
$(\lambda_{\mathrm{cm}}, \lambda_{\mathrm{fm}}, \lambda_{\mathrm{adv}})$ & (0.25, 25.0,  4.0) \\ 
\bottomrule
\\
\toprule
\multicolumn{2}{l}{\bf Transformer-NMT training condition}  \\
\midrule
\# Encoder blocks & 6 \\
\# Decoder blocks & 6 \\
Feed-forward units & 2,048 \\
Attention dimension & 256 \\
\# Attention heads & 4 \\
Dropout-rate & 0.1 \\
Batch size & 96 (in average) \\
Optimizer & Noam~\cite{vaswani2017attention} \\
\# Warmup steps & 8,000 \\
Gradient clip & 5 \\
\# Epochs & 2,000 \\
Label smoothing weith & 0.1 \\
\bottomrule
\end{tabular}
}}
\vspace{-5mm}
\label{tb:ex_cond_1}
\end{center}
\end{table}
While we used a fixed batch size with randomly cropped speech for the VQ-VAE model, using a dynamic batch making depending on the length of each sequence for the Transformer-NMT model.
All of the models were trained with a single GPU (Titan V) with open-source E2E speech processing toolkit ESPnet~\cite{watanabe2018espnet}.

\subsection{Objective evaluation results}
We conducted the objective evaluation using the ASR-based objective measure character error rate~(CER) and the token error rate (TER) of the NMT models.
As the ASR model, we used the Transformer-ASR model trained on the corpus of spontaneous Japanese (CSJ)~\cite{maekawa2003corpus}.
The objective evaluation result is shown in Table~\ref{tb:obj_result_1}\footnote{Note that ASR result includes Hiragana, Katakana, and Kanji. It contains many homophonic words (e.g., \Ja{砂糖} and \Ja{佐藤}) and transcription mismatch (e.g., \Ja{それほどに} and \Ja{それ程に}).}, where ``Beam'' represents the beam-size of beam-search in decoding.

\begin{table}[t]
\begin{center}
\vspace{0mm}
\vspace{0mm}
\caption{\it Character error rates calculated by the ASR model and token error rates of the NMT models.}
\vspace{-2mm}
\scalebox{0.9}{\renewcommand\arraystretch{0.9}{%
\begin{tabular}{lcccc}
\toprule
Method                   & CER [\%] & TER [\%] \\
\midrule
Baseline                 & 15.1     & -        \\
Reconst (DSF256)         & 22.9     & -        \\
Proposed (DSF256, Raw)   & 23.8     & 87.2     \\
\ $+$ Beam=3             & 23.1     & 87.2     \\
\ $+$ Beam=5             & 23.5     & 87.4     \\
\ $+$ Beam=10            & 22.9     & 87.3     \\
Proposed (DSF256, SW256) & 24.7     & 92.4     \\
\ $+$ Beam=3             & 23.8     & 92.3     \\
\ $+$ Beam=5             & 22.9     & 92.2     \\
\ $+$ Beam=10            & 23.5     & 92.2     \\
Reconst (DSF128)         & 14.8     & -        \\
Proposed (DSF128, Raw)   & 20.8     & 91.5     \\
\ $+$ Beam=3             & 21.3     & 92.0     \\
\ $+$ Beam=5             & 21.9     & 92.4     \\
\ $+$ Beam=10            & 21.2     & 92.6     \\
Proposed (DSF128, SW256) & 18.6     & 93.3     \\
\ $+$ Beam=3             & 18.2     & 93.0     \\
\ $+$ Beam=5             & 19.1     & 93.0     \\
\ $+$ Beam=10            & 19.0     & 93.2     \\
Proposed (DSF128, SW512) & 19.5     & 94.8     \\
\ $+$ Beam=3             & 19.1     & 94.8     \\
\ $+$ Beam=5             & 19.6     & 94.9     \\
\ $+$ Beam=10            & 19.3     & 94.8     \\
\midrule
Target                   & 12.1     & -        \\
\bottomrule
\end{tabular}
}}
\vspace{-5mm}
\label{tb:obj_result_1}
\end{center}
\end{table}

First, we focus on the CER result in Table~\ref{tb:obj_result_1}. A comparison between the different DSFs shows that the reconstruction with the small DSF achieved the performance comparable to the baseline while the use of the large DSF greatly affected the intelligibility. 
From audio samples available in \cite{audio_samples},  the model with a large DSF generated slurred speech while keeping high signal-to-noise (SN) ratio.
The authors highly recommend listening to the samples to understand the difference.
Thus, there was a trade-off between the size of DSF and speech articulation.
From a comparison between the target types, the use of the subword was effective, especially in the case of the small DSF. 
This is because the same discrete symbol repeatedly appeared in the sequence due to high time resolution, and the subword can summarize these successive symbols.
Note that since the use of the subword made the length of the target sequence smaller, it can also reduce the training time.
However, the use of a large number of subword units made the target symbols sparse, decreasing the performance.
Therefore, we need to tune the number of subword units according to the size of the training data.
Focusing on the effectiveness of beam-search, it led to a slight improvement of the performance, but a large beam-size did not always bring the improvement.

Next, we focus on the TER result of the NMT models in Table~\ref{tb:obj_result_1}.
Interestingly, the predicted sequence is totally different from the ground-truth, and we could not find the meaningful correlation between the CER of the ASR model and the TER of the NMT model.
The training process also tended to be over-fitting in the early stage.
One of the possible reasons is that speech included various speaking styles or intonations, even for the same words.
Therefore, the conversion from the text to the sequence of discrete symbols became a one-to-many problem. 

Finally, we investigated the performance of shallow fusion with the LM for the subword sequence of the discrete symbols (VQ-LM).
We built three long short-term memory (LSTM)-based VQ-LMs with 1,024 units and a different number of layers (1, 2, and 4) for the subword sequence (DSF128, SW256).
The training curve showed the same tendency as the NMT models, which tended to overfit in the early stage, and the deeper model brought lower training perplexity (10.6, 7.6, and 5.3, respectively).
To check the effectiveness of the fusion, we used the best perplexity VQ-LM model and changed the weight for its score in decoding from 0.1 to 0.3 while fixing beam-size to 3.
The result is shown in Table~\ref{tb:obj_result_2}.
From the results, we could not confirm the improvement.
One of the reasons is that the training data was the same for NMT models and VQ-LMs, and the amount of training data was relatively small while the length of each sequence was long.
We need to investigate the case where the training data of VQ-LM is much bigger than NMT models; in other words, we have many untranscribed utterances.

\subsection{Subjective evaluation results}
We conducted a subjective evaluation using mean opinion score (MOS) on naturalness.
We used the ``VOICE ACTRESS'' subset in the JSUT corpus (= 100 utterances) for the subjective evaluation.
The number of subjects is 45, and that of evaluation samples per each subject is 160 (= 20 samples $\times$ 8 models).
Each subject rated the naturalness of each sample on a 5-point scale: 5 for excellent, 4 for good, 3 for fair, 2 for poor, and 1 for bad.
We instructed subjects to work in a quiet room and use headphones.
We used the Likert single stimulus test available in WebMUSHRA~\cite{schoeffler2018webmushra}.

\begin{table}[t]
\begin{center}
\vspace{0mm}
\vspace{0mm}
\caption{\it Effectiveness of shallow fusion with the VQ-LM.}
\vspace{-2mm}
\scalebox{0.9}{\renewcommand\arraystretch{0.9}{%
\begin{tabular}{cccc}
\toprule
\multirow{1}{*}{LM weight} & CER [\%] & TER [\%]\\
\midrule
0.1                        & 18.4     & 93.2 \\
0.2                        & 21.0     & 93.8 \\
0.3                        & 21.4     & 94.2 \\
\midrule
0.0                          & 18.2     & 93.0 \\
\bottomrule
\end{tabular}
}}
\vspace{-3mm}
\label{tb:obj_result_2}
\end{center}
\end{table}

The subjective result is shown in Table~\ref{tb:sub_result_1}.
A comparison between the different DSFs shows that the small DSF led to better naturalness than the large one in the reconstruction condition.
In the case of the large DSF, both raw and subword models are almost the same naturalness, comparable to the reconstruction condition.
However, in the case of the small DSF, there is a large difference between raw and subword models.
While the subword model achieved the performance comparable to the reconstruction condition, the raw model is worse than the reconstruction.
The reason for the difference is that due to the lengthened target sequence, the raw model with the small DSF sometimes failed to generate speech, leading long pause and word deletions.
Hence, the use of the subword unit is effective, especially in the case of the small DSF.
Compared to the other model, our best model outperformed the baseline with a significant difference (significance level of 5 \%). 

The results represent that the VQ-VAE model determined the upper bound of the performance, and there was less gap between the reconstruction and synthesis conditions than the conventional E2E-TTS models.
Also, while the smaller DSF led to better reconstruction performance, but the smaller one made the training of NMT models difficult. 
Therefore, if we can improve the reconstruction performance of the VQ-VAE model while keeping a reasonable DSF size, it is expected that the quality of the proposed method is further improved.
\begin{table}[t]
\begin{center}
\vspace{0mm}
\vspace{-1mm}
\caption{\it Mean opinion score on naturalness, where CI represents confidence interval. The beam-size is fixed to 1.}
\vspace{-2mm}
\scalebox{0.9}{\renewcommand\arraystretch{0.9}{%
\begin{tabular}{lc}
\toprule
Method                  & MOS $\pm$ 95\% CI \\
\midrule
Baseline                 & 3.48 $\pm$ 0.08   \\
Reconstruction (DSF256) & 3.36 $\pm$ 0.07   \\
Proposed (DSF256, Raw)       & 3.25 $\pm$ 0.07   \\
Proposed (DSF256, SW256)   & 3.27 $\pm$ 0.07   \\
Reconstruction (DSF128) & 3.99 $\pm$ 0.06   \\
Proposed (DSF128, Raw)       & 3.39 $\pm$ 0.08   \\
Proposed (DSF128, SW256)   & 3.93 $\pm$ 0.06   \\
\midrule
Target                  & 4.32 $\pm$ 0.05   \\
\bottomrule
\end{tabular}
}}
\vspace{-5mm}
\label{tb:sub_result_1}
\end{center}
\end{table}

\section{Conclusions}
This paper proposed a novel E2E-TTS framework consisting of the VQ-VAE and NMT models.
The experimental evaluation with the JSUT corpus showed that 
1) the proposed model outperforms the conventional Transformer-TTS with Parallel WaveGAN in naturalness, achieving the performance comparable to the reconstruction condition, 
2) the use of subword unit is effective, especially in the case of the small downsampling factor, 
and 3) there is a trade-off between the resolution of discrete symbols and speech articulation.

In future work, we will consider the attention constraint for Transformer-NMT to make the generation more stable, extend this framework to a multi-speaker model, apply to the large-scale corpus to clarify the effectiveness of VQ-LMs, and make it fully-non-autoregressive by using connectionist temporal classification~\cite{graves2006connectionist}.

\clearpage
\section{References}
{
\setstretch{0.9}
\printbibliography
}
\end{document}